\documentclass[letterpaper,journal]{IEEEtran}
\usepackage[T1]{fontenc}
\usepackage{amsmath,amssymb}
\usepackage{graphicx}
\usepackage{cite}
\usepackage{xurl}
\usepackage[hidelinks]{hyperref}
\hypersetup{pdftitle={AcrossDEX 1.0: From Human Motion Pretraining to Bimanual Dexterous Manipulation across Hand Embodiments},
 pdfauthor={Xiaoyu Yang and Nan Wu}}
\newcommand{\SO}{\mathrm{SO}(3)}
\begin{document}
\title{AcrossDEX 1.0: From Human Motion Pretraining to Bimanual Dexterous Manipulation across Hand Embodiments}
\author{Xiaoyu~Yang and Nan~Wu%
\thanks{Corresponding author: Nan Wu (e-mail: \href{mailto:across2026@163.com}{across2026@163.com}).}}
\maketitle

\begin{abstract}
Bimanual dexterous manipulation requires continuous coordination of wrist motion, finger contact, and object motion. Human videos provide rich manipulation experience. Using this experience for robot control requires aligning human poses with robot commands and accounting for embodiment-specific kinematics and contact dynamics. We present AcrossDEX 1.0, a learning system that connects human motion pretraining, simulation-based expert learning, robot demonstration collection, and visual closed-loop execution. Pretraining on HOT3D Aria predicts one second of bimanual motion from five images and thirty annotated hand states over the preceding second. Robot adaptation preserves the visual encoder, temporal backbone, and wrist parameters. Shared wrist representations, separate finger fields, and validity masks support both six-channel Inspire hands and sixteen-joint Allegro hands. The robot policy predicts sixty control targets from frontal images, command history, and task conditions. It executes a prefix and replans from updated observations. We compare direct regression, diffusion, and flow matching on lid opening and return and closed-lid object reorientation. Direct regression achieves success rates of 40.6\% for opening and return, 100.0\% for Inspire reorientation, and 93.3\% for Allegro reorientation. Flow matching achieves 90.0\% on Allegro reorientation. Paired comparisons further show that constraining replanning with the previous plan improves the continuous execution of generative policies. These results demonstrate a common adaptation pipeline for two hand embodiments and highlight the roles of contact coordination and action-chunk continuity.
\end{abstract}


\section{Introduction}
\IEEEPARstart{B}{imanual} dexterous manipulation changes an object's state through sustained coordination between two hands. During lid opening, one hand stabilizes the box while the other moves the lid. During object reorientation, both hands adjust their actions while maintaining contact support. These tasks couple wrist motion, finger articulation, and contact dynamics. A policy must coordinate motion amplitude, timing, and the roles of the two hands. Human manipulation videos contain rich visual and motion information about this coordination. Robot actuation, mechanical coupling, and available observations determine how this information can inform control. A six-channel coupled hand and a sixteen-joint hand have different control spaces. The same human motion therefore requires distinct robot interfaces. Learning bimanual manipulation from human videos calls for a coherent pipeline that connects motion representations, robot control targets, and visual feedback.

Research on human data has progressed from visual representation learning to motion modeling and robot skill acquisition. HOT3D provides egocentric views of hand--object interactions with three-dimensional annotations~\cite{hot3d}. ARCTIC captures bimanual interactions with articulated objects~\cite{arctic}, while EgoDex studies large-scale egocentric data for dexterous manipulation~\cite{egodex}. These datasets connect object appearance, hand pose, and their temporal relationships. In robot learning, R3M extracts transferable visual representations from human videos~\cite{r3m}. VideoDex uses human motion priors to guide robot behavior~\cite{videodex}, and DexMV converts human manipulation videos into demonstrations for imitation learning~\cite{dexmv}. AcrossDEX follows this direction through explicit prediction of future bimanual trajectories. It jointly encodes images and pose history during pretraining. The resulting parameters support visual perception, temporal modeling, and wrist prediction before robot data introduce specific control targets.

Transferring human motion to robot control also requires demonstrations that are compatible with contact dynamics. Early studies showed that reinforcement learning can acquire dexterous contact skills from demonstrations and large-scale simulation~\cite{dapg,openaihand}. Subsequent work has explored several routes for collecting demonstrations and retargeting motion. DexCap provides a portable motion-capture system for dexterous manipulation~\cite{dexcap}. ViViDex connects human-video-guided skill learning in simulation with visual policy training~\cite{vividex}. DexMachina trains bimanual experts through functional retargeting and curriculum learning~\cite{dexmachina}. DexImit generates bimanual robot data from monocular human videos~\cite{deximit}. AcrossDEX adopts DexMachina's demonstration assets, rewards, and training procedure to obtain simulation experts. It records the experts' final issued control targets as robot supervision. Human pretraining and simulation experts thus play complementary roles. Pretraining learns visual and motion representations, while experts provide embodiment-specific experience of physical interaction. Robot post-training brings these sources together.

Connecting data sources and hand embodiments requires both semantic alignment and temporal organization. Open X-Embodiment demonstrates the value of organizing data and learning across robots~\cite{oxe}. Being-H0.5 studies human-centric cross-embodiment learning with a unified action representation~\cite{being}. At execution time, ACT uses action chunks for bimanual control~\cite{act}, Diffusion Policy models action distributions through conditional diffusion~\cite{diffusionpolicy}, and $\pi_0$ generates control sequences with a flow model~\cite{pi0}. These policies represent future motion as action chunks and respond to new observations through receding-horizon replanning. Continuity between successive chunks is therefore part of control quality. RTC constrains new chunks with the previous plan and uses asynchronous execution to handle inference delays~\cite{rtc}. ACG improves action coherence through attention-based inference guidance~\cite{acg}. AGRoL studies time conditioning and motion smoothness in human motion generation~\cite{agrol}. These studies motivate two design choices in AcrossDEX: shared wrist geometry with separate finger fields, and coordinated execution of successive action plans.

AcrossDEX 1.0 integrates human motion pretraining, simulation expert data, and single-camera closed-loop control into one pipeline. It predicts one second of future motion from one second of images and hand history. Direct regression, diffusion, and flow matching use the same data and backbone configuration. Robot adaptation retains pretrained visual, temporal, and wrist parameters while introducing finger fields for Inspire and Allegro hands. Thirty human prediction steps become sixty robot command steps through a shared physical time horizon. We train separate policies for each task and hand embodiment. Lid opening and return and closed-lid reorientation evaluate the complete pipeline. Paired action-chunk comparisons examine the effect of continuity during execution. Together, these components connect human motion modeling, cross-embodiment control interfaces, and physical closed-loop evaluation.

\section{Method}
\label{sec:method}
\subsection{Overview and Problem Formulation}
\begin{figure*}[!t]
\centering
\includegraphics[width=\textwidth]{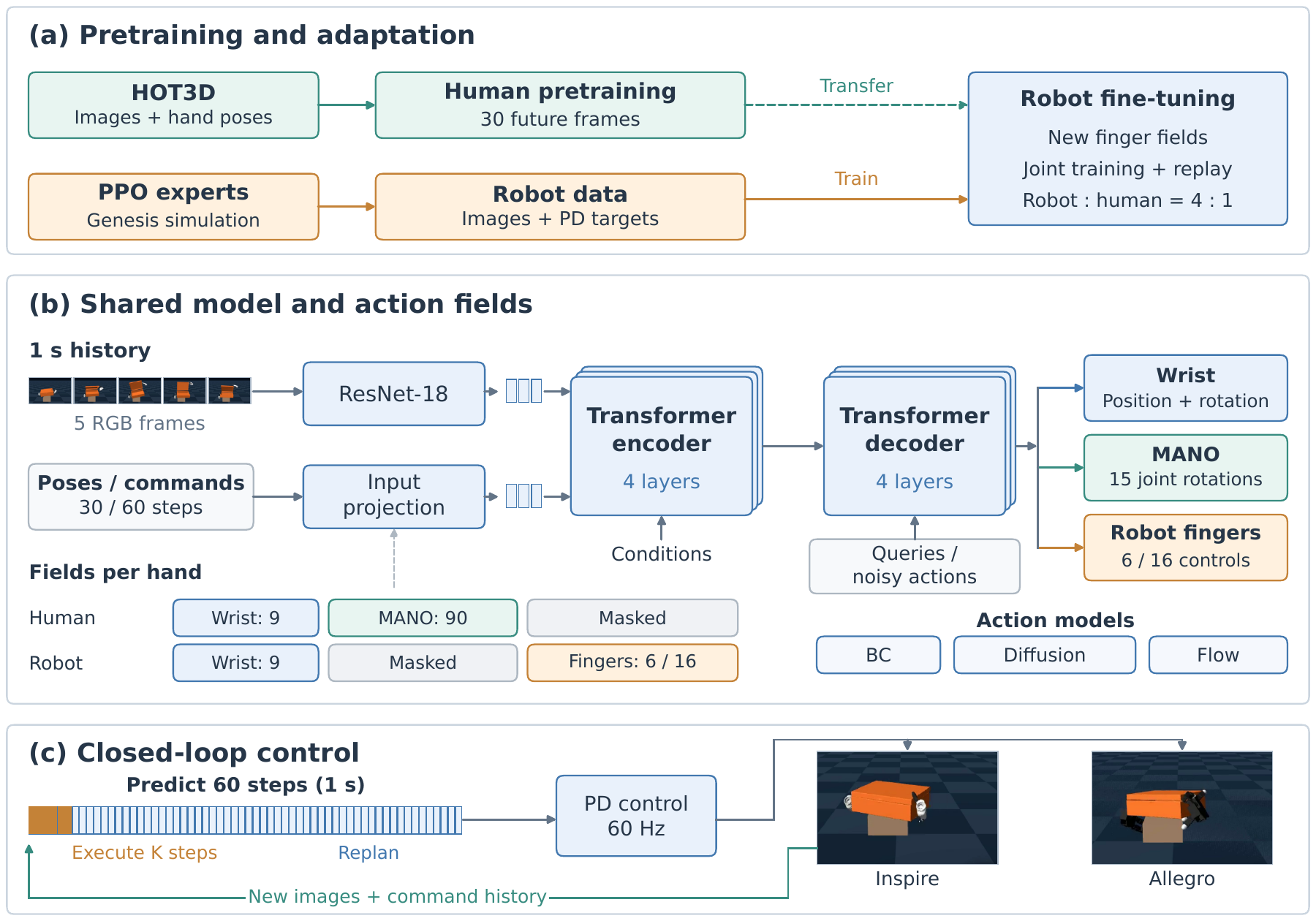}
\caption{Overview of AcrossDEX 1.0. (a) Human pretraining and robot expert data support adaptation with human replay. (b) A shared Transformer processes images and motion history. Masks select human or robot finger fields. Each method, task, and hand uses a separately trained policy. (c) The robot policy predicts 60 commands, executes the first $K$ steps, and updates its history. Blue, green, and orange denote shared, human, and robot components. Gray fields are masked.}
\label{fig:pipeline}
\end{figure*}

Figure~\ref{fig:pipeline} summarizes the three connected stages of AcrossDEX: human trajectory pretraining, robot expert learning, and visual policy post-training. The human model predicts motion from images and annotated poses. Simulation experts learn physical interaction from privileged states and generate paired images and control targets. Post-training inherits the human model's shared parameters and learns robot command sequences through embodiment-specific interfaces. The resulting student policy controls both hands from a single camera, issued-command history, and task conditions. This design assigns explicit roles to pretraining targets, demonstration labels, and deployment inputs. Shared parameters and action encoding connect the stages.

Let $t$ denote the robot control step, $I_t$ the frontal image, $u_t$ the final issued bimanual command, and $g$ the task condition. The policy predicts one second of commands from the preceding second of observations and commands:
\begin{equation}
 \begin{aligned}
 U_t&=\pi_\theta\!\left(\mathcal I_t,u_{t-60:t-1},u_{t-1}^{\mathrm{anchor}},c_t\right),\\
 U_t&=(\hat u_t,\ldots,\hat u_{t+59}).
 \end{aligned}
 \label{eq:policy}
\end{equation}
Here, $\mathcal I_t$ is the sampled image history and $c_t=(g,t,\mathcal C,\mathcal M)$ collects task, time, calibration, and validity information. The calibration condition is $\mathcal C$, while $\mathcal M$ marks valid time steps and action fields. The command anchor $u_{t-1}^{\mathrm{anchor}}$ defines the reference pose for future wrist motion. At $t=0$, it is the initial PD target. Short histories are padded and masked. The policy executes the first $K$ predicted steps, acquires a new image, and updates its command history. Our main setting uses $K=6$. Opening experiments also compare $K=10,30,60$. The prediction horizon determines the planned motion, while the execution horizon determines the frequency of feedback updates.

\subsection{Human History and Future Motion Representation}
Human pretraining uses HOT3D \texttt{train\_aria}. Thirty hand states, including the current state, form the history. The target contains the next thirty bimanual states. At approximately $30\,$Hz, each sequence spans one second. Five images are sampled at history indices $[0,7,14,22,29]$. Sparse visual sampling captures scene changes, while the complete pose sequence preserves motion information. Left and right hands occupy fixed slots, with masks marking valid annotations. Calibrated fisheye images are converted to pinhole views, rotated upright, and resized to $224\times224$. Three-dimensional annotations undergo the corresponding transformation into a fixed camera frame at the observation time. This separates head-mounted camera motion from hand motion. Let $F$ denote the fixed frame, $C_0$ the current world-to-original-camera transform, and $U$ the rigid transform used to rotate the image upright. Then
\begin{equation}
 \begin{aligned}
 T_{F\leftarrow W}&=UC_0,\\
 \begin{bmatrix}p_k^F\\1\end{bmatrix}
 &=T_{F\leftarrow W}\begin{bmatrix}p_k^W\\1\end{bmatrix},\\
 R_k^F&=R_{F\leftarrow W}R_k^W.
 \end{aligned}
 \label{eq:camera}
\end{equation}
The rotation component is $R_{F\leftarrow W}$. The same transform applies to the history and future within each window. Wrist trajectories therefore share one reference frame. Future motion is then expressed relative to the current hand pose. Let $p_k$ be the wrist position, $R_k\in\SO$ its rotation, and $Q_{k,j}\in\SO$ the local rotation of MANO joint $j$. Omitting the superscript $F$, we define
\begin{equation}
 \begin{aligned}
 \Delta p_k&=p_k-p_0, & \Delta R_k&=R_0^\top R_k,\\
 \Delta Q_{k,j}&=Q_{0,j}^\top Q_{k,j}.
 \end{aligned}
 \label{eq:relative}
\end{equation}
Every future state uses the same anchor. Absolute rotations are recovered through matrix composition, such as $R_k=R_0\Delta R_k$. Historical wrist poses use absolute coordinates in the fixed reference frame. Local finger rotations follow MANO's joint definitions~\cite{mano}. The fifteen finger rotations are read as a complete 45-dimensional axis-angle vector. They are converted to continuous six-dimensional rotations for the network~\cite{rotation6d}. Each hand therefore has $3+6+15\times6=99$ fields, and the future tensor has shape $30\times2\times99$. MANO shape parameters support geometric reconstruction and constraint evaluation. Normalization statistics, feasible joint-angle ranges, and related priors use training data. This representation separates wrist motion from local finger articulation while preserving a common anchor and rotation reconstruction rule.

\subsection{Shared Backbone and Motion Prediction Objectives}
The three prediction methods share a visual and temporal backbone. This configuration compares action models under the same conditioning information. The visual encoder is an ImageNet-pretrained ResNet-18~\cite{resnet}. We freeze its stem, layer1, and layer2, fix batch-normalization statistics, and fine-tune the remaining layers. Each image retains $7\times7$ spatial features. The five images yield 245 visual tokens. They enter a Transformer alongside sixty bimanual state tokens and five camera-pose tokens~\cite{transformer}. The Transformer has four encoder layers and four decoder layers. It uses a feature width of 256, eight attention heads, and a feedforward width of 1024. Time-indexed future queries produce the complete action chunk. Each predicted time step can therefore use both visual context and historical motion.

We compare direct regression, conditional diffusion, and flow matching on this backbone. BC denotes the direct-regression implementation of behavior cloning. Given conditions $c$, it predicts the normalized trajectory $x$ through $f_\theta(c)$ and minimizes a masked mean squared error. Diffusion uses the DDPM noising process~\cite{ddpm} with velocity parameterization. A normalized trajectory and Gaussian noise $\epsilon$ form the noisy input:
\begin{equation}
 \begin{aligned}
 x_n&=a_nx+b_n\epsilon,\\
 v_n^{*}&=a_n\epsilon-b_nx,\qquad a_n^2+b_n^2=1.
 \end{aligned}
 \label{eq:diffusion}
\end{equation}
The network learns to predict $v_n^{*}$. Its clean trajectory estimate is $\hat x=a_nx_n-b_nv_\theta$. Training uses 1,000 diffusion steps and a zero-terminal-SNR schedule. Inference uses fifty DDIM steps~\cite{ddim} with $\eta=0$. A sampling seed determines the initial noise. Flow matching~\cite{flowmatching} also generates trajectories from Gaussian noise, but its training path directly connects noise and data:
\begin{equation}
 \begin{aligned}
 x_\tau&=(1-\tau)\epsilon+\tau x,\qquad v^{*}=x-\epsilon,\\
 \hat x&=x_\tau+(1-\tau)v_\theta(x_\tau,\tau,c).
 \end{aligned}
 \label{eq:flow}
\end{equation}
Flow learns this path by regressing its velocity field. Inference uses thirty-two Heun integration steps. The three methods thus share inputs and backbone capacity while retaining their own targets and inference procedures. Human training balances translation, wrist rotation, and finger rotation by averaging the mean squared error within each group. The three group losses receive equal weight. Let $i$ index samples, future times, and hands. Let $m_i$ denote annotation validity and $d_g$ the dimension of field group $g$. The objective is
\begin{equation}
 \begin{aligned}
 \mathcal L_{\mathrm h}
 ={}&\frac{1}{3}\sum_{g\in\{p,R,Q\}}
 \frac{\sum_i m_i\,\|\hat y_{i,g}-y_{i,g}^{*}\|_2^2/d_g}
      {\max(1,\sum_i m_i)}\\
 &+\lambda_{\mathrm{BMC}}\mathcal L_{\mathrm{BMC}}(\hat x).
 \end{aligned}
 \label{eq:humanloss}
\end{equation}
We use $d_p=3$, $d_R=6$, $d_Q=90$, and $\lambda_{\mathrm{BMC}}=0.01$. The target $y^*$ is the normalized trajectory for BC, diffusion velocity for Diffusion, and flow velocity for Flow. The corresponding network output is $\hat y$. Inspired by biomechanical constraints (BMC)~\cite{bmc}, we estimate feasible flexion and abduction regions from left- and right-hand training annotations. These constraints act on reconstructed clean hand poses. BC uses its direct prediction, while the generative methods use their clean trajectory estimates. Trajectory supervision models motion over time. BMC complements it by encouraging geometrically plausible hand configurations.

\subsection{Simulation Experts and Robot Data}
Robot experts use DexMachina's functional retargeting framework~\cite{dexmachina} and are trained with PPO~\cite{ppo} in Genesis. They observe simulator states, contacts, and reference demonstrations. Rewards and curricula follow the original definitions for reference object motion, hand actions, and contact targets. Progressively reduced object assistance supports skill acquisition. Final evaluation and formal data collection use zero object assistance. Object motion then results from hand contact and simulation dynamics. Both embodiments use floating wrists, and the box initially rests on a fixed support. Each Inspire hand has six active finger controls, with mechanical coupling defined by the asset. Each Allegro hand has sixteen controlled finger joints. Adding six wrist position and orientation controls per hand gives 24-dimensional and 44-dimensional bimanual commands, respectively.

Data collection records the controller's final targets to align expert outputs with student supervision. The expert adds policy corrections to the reference wrist trajectory and generates finger targets. After combining and clipping these values, the collector stores the commands actually sent to the PD controller. It also records frontal RGB images, timestamps, task targets, trajectory boundaries, and diagnostic states. Reference motion thus supports expert learning and data generation. The student learns a direct mapping from available observations to final wrist and finger commands. Images and command sequences provide the main post-training data. Simulator states and contact records support data checks and physical execution evaluation.

\subsection{Embodiment Fields and Parameter Transfer}
Human poses and robot commands share wrist geometry but describe different finger motion spaces. MANO rotations represent human articulation, while robot finger fields represent embodiment-specific joint targets. Robot history contains previously issued commands. Following the semantic alignment principle of Being-H0.5~\cite{being}, we retain the human wrist and MANO fields and append separate robot finger fields. Validity masks select the components used in each domain. The per-hand width is $99+6=105$ for Inspire and $99+16=115$ for Allegro. Human samples activate three translation, six wrist rotation, and ninety MANO fields. Robot samples activate the nine wrist fields and their own finger fields. Original input and output parameters remain associated with the human fields. Post-training learns the appended parameters and domain conditions. The visual encoder, temporal backbone, and wrist prediction parameters are inherited from the human model. Let $y\in\mathbb R^{99+d_e}$ be the normalized input after learnable scale and offset calibration of the robot wrist. Let $m_e$ be the embodiment mask and $d_e\in\{6,16\}$ the robot finger dimension. The expanded projection is
\begin{equation}
 \begin{aligned}
 z_e={}&W_{\mathrm h}(m_e\odot y)_{1:99}\\
      &+W_e(m_e\odot y)_{100:99+d_e}+b_{\mathrm h}.
 \end{aligned}
 \label{eq:interface}
\end{equation}
The inherited parameters are $W_{\mathrm h}$ and $b_{\mathrm h}$, while $W_e$ is the new finger projection. Human samples set the appended fields to zero. Robot samples set the MANO finger fields to zero. The output layer follows the same principle: it retains existing rows and appends robot finger rows. Robot wrist targets are encoded relative to the current issued-command anchor using the geometry in~\eqref{eq:relative}. Finger targets are absolute joint angles in radians. Decoding transforms wrist poses into the simulation world frame, selects a continuous orientation branch, and applies joint limits. This encoding and decoding process connects shared motion representations to executable robot commands.

The temporal interface links both training stages through physical duration. Robot data are recorded at $60\,$Hz. Inputs contain the sixty commands issued before the current step. Image offsets are $[-59,-44,-30,-15,0]$. Outputs cover the current command and the following fifty-nine commands. The human model's thirty future queries are interpolated to sixty according to physical time, preserving a one-second prediction horizon. Additional conditions include camera calibration, task targets, duration, and current time. Reorientation targets specify object position, orientation, and lid angle. Opening targets specify the opening angle, return angle, and task duration. Frontal images and issued commands form the student's observation history. Actual joint states, object states, and contacts support expert learning and diagnostic scoring.

Post-training consists of interface warm-up and joint fine-tuning. Each method, task, and embodiment starts from its corresponding human model trained for five hundred epochs. Stage A performs 1,024 updates with robot supervision to learn the new interfaces. Stage B jointly updates the shared backbone and new interfaces. Human data are replayed at a robot-to-human batch ratio of $4:1$, so shared parameters receive both forms of motion supervision. Each logical epoch contains 32,768 robot windows and 8,192 human windows. With a batch size of 128, this gives 320 updates. Robot losses use valid time steps and active action fields. Human batches retain their original trajectory objective and BMC term. Let $\mathcal L_{\mathrm r}$ be the mean squared error over valid robot action scalars. With robot and human distributions $\mathcal D_{\mathrm r}$ and $\mathcal D_{\mathrm h}$, the expected Stage B objective is
\begin{equation}
 \begin{aligned}
 \mathcal J_B(\theta)
 ={}&\frac45\,\mathbb E_{\mathcal D_{\mathrm r}}
       [\mathcal L_{\mathrm r}(\theta)]\\
   &+\frac15\,\mathbb E_{\mathcal D_{\mathrm h}}
       [\mathcal L_{\mathrm h}(\theta)].
 \end{aligned}
 \label{eq:adaptation}
\end{equation}
The implementation alternates four robot batches with one human batch. The weights in~\eqref{eq:adaptation} express this sampling frequency. Learning the new interface before jointly updating shared representations supports adaptation to robot targets while maintaining the human trajectory task.

\subsection{Receding-Horizon Execution and Chunk Continuity}
The policy updates its future plan after each new observation. This allows subsequent commands to respond to object motion. BC produces a deterministic prediction for a given input. Diffusion and Flow generate chunks from noise, which can introduce additional variation between successive plans. Sustained contact requires smooth changes within each chunk and coordinated transitions between chunks. We use the soft-prefix principle of RTC~\cite{rtc}. The unexecuted portion of the previous plan is decoded into commands and re-encoded relative to the current anchor. This gives a prefix target $z$ in the new plan's coordinates. Let $\hat x(x_\tau)$ denote the current clean trajectory estimate and $w_i$ the prefix weight. The guidance energy and corrected velocity are
\begin{equation}
 \begin{aligned}
 E&=\frac12\sum_i w_i\left\|\hat x_i-z_i\right\|_2^2,\\
 \widetilde v&=v_\theta-\gamma(\tau)\nabla_{x_\tau}E.
 \end{aligned}
 \label{eq:rtc}
\end{equation}
Weights decay along the remaining prefix and are zero outside it. Guidance therefore acts on the overlap between successive plans. Gradients use the full vector--Jacobian product, with guidance gain capped at five. Flow applies the term on its linear path. Diffusion first converts its velocity parameterization to an equivalent linear path, applies guidance, and converts back. Closed-loop evaluation uses synchronous execution. Simulation pauses during inference and advances at $60\,$Hz after commands become available. RTC coordinates adjacent chunks in this setting. The main Flow and Diffusion results use RTC, while BC uses direct regression. Opening experiments also compare ACG~\cite{acg} and its combination with RTC.

\section{Experimental Setup}
\subsection{Data and Training Scale}
Human data processing covers 124 original HOT3D records and 227,400 frames. The 99 records used for optimization contain 195,300 frames, equivalent to about 108.5 minutes at $30\,$Hz. Data are organized by complete recordings, and each window inherits its recording's assignment. Normalization and hand-shape priors are estimated from training data. Window construction checks recording boundaries and timestamp gaps. Each logical epoch samples 32,768 distinct starts from 163,369 valid positions. Starts are resampled in the next epoch. This fixed window budget covers different temporal locations, while windows from different starts may overlap. Fixed offline evaluation windows use an eight-frame stride. All three human models are trained for five hundred epochs and each reads 16,384,000 windows. BC uses batch size 32 for the first twenty-three epochs and 128 thereafter, for 145,664 updates. Diffusion and Flow use batch size 128 throughout, with 128,000 updates each.

Robot data cover Inspire lid opening and return, Inspire closed-lid reorientation, and Allegro closed-lid reorientation. The opening dataset retains 1,770 of 2,000 trajectories whose peak lid angle reaches at least $90^\circ$. Inspire reorientation retains 298 of 300 trajectories after quality filtering and manual inspection. Allegro reorientation uses 293 trajectories accepted during collection. The numbers used for optimization are 1,413, 239, and 233, respectively. Stage B runs for three hundred epochs on opening and one hundred epochs on each reorientation task. Reorientation demonstrations use an independently cropped closed-lid segment. The reference rotates by approximately $63.96^\circ$ over $0.67\,$s. The model retains sixty history and prediction slots, using padding and validity masks for this short segment. History accumulates during execution. Under the fixed image offsets, each reorientation window contains at most three valid historical images. Valid future commands extend to the segment's end.

Collection applies small perturbations around the same box asset and reference motion. These cover wrist position and orientation, finger angles, friction, camera pose, and lighting. The demonstrations therefore include local variation in initial conditions. Each dataset is checked for zero physical object assistance. Sixteen same-batch trajectories are replayed to verify zero discrepancy between stored commands and replay targets. Student evaluation inherits the original scene slots used during collection. Other slots replay their recorded PD targets, preserving the rendering and simulation context. Complete image, command, and state sequences support data alignment, control replay, and task scoring.

\subsection{Training and Inference Details}
Training uses separate learning rates for pretrained modules and new robot interfaces. Human pretraining uses AdamW. Learning rates are $10^{-4}$ for the main network and $10^{-5}$ for the visual encoder, with weight decay $10^{-4}$ and gradient clipping at one. Robot post-training uses learning rates of $10^{-4}$, $10^{-5}$, and $10^{-6}$ for new, shared, and visual parameters, respectively. Microbatch and effective batch sizes are both 128. Stages A and B initialize separate optimizer states. Including warm-up, each reorientation model receives 33,024 updates and each opening model receives 97,024 updates. Simulation experts use 4,096 parallel environments and collect thirty-two steps per update. PPO uses a minibatch of 32,768 and five optimization epochs. Inspire experts receive five thousand updates, corresponding to 655,360,000 interactions. The Allegro expert receives fifteen hundred updates, corresponding to 196,608,000 interactions. Each training run uses one GPU with 40\,GB of memory. Visual policy inference follows the samplers and execution scheme described in Section~\ref{sec:method}.

\subsection{Evaluation and Result Selection}
\label{sec:evaluation}
Evaluation uses physical simulation trajectories driven by the student policies. Task success measures completion of the full operation. The opening task requires approach, opening, and lid return while maintaining box height. Reorientation requires the target orientation with a closed lid and sufficient object height. Let $\theta_t$ be the lid angle and $z_t$ the box root height. Let $R_T$ and $R_{\mathrm{goal}}$ denote the final and target object orientations. The geodesic rotation error is
\begin{equation}
 d_R(\hat R,R)=\arccos\!\left(\operatorname{clip}\left(
 \frac{\operatorname{tr}(\hat R^\top R)-1}{2},-1,1\right)\right).
 \label{eq:geodesic}
\end{equation}
Let $T$ be the number of executed steps. The indicator $\mathbf1[\cdot]$ equals one when all enclosed conditions hold. The task criteria and success rate are
\begin{equation}
 \begin{aligned}
 S_{\mathrm{open}}=\mathbf1\big[&T=200,\ \max_t\theta_t\ge90^\circ,\\
                         &\theta_T\le25^\circ,\ \min_t z_t\ge0.9\,\mathrm m\big],\\
 S_{\mathrm{turn}}=\mathbf1\big[&T=40,\ d_R(R_T,R_{\mathrm{goal}})\le10^\circ,\\
                         &\max_t|\theta_t|\le5^\circ,\ \min_t z_t\ge0.9\,\mathrm m\big],\\
 \mathrm{SR}={}&100\%\times\widehat{\mathbb E}[S].
 \end{aligned}
 \label{eq:success}
\end{equation}
Here, $\widehat{\mathbb E}$ is the empirical average over evaluation trajectories. Opening criteria combine peak angle, final return angle, and minimum height. Reorientation criteria combine final orientation, lid angle throughout the motion, and minimum height. Success therefore reflects a complete physical operation. For each configuration, the main results report the model and corresponding run with the highest success rate over a complete evaluation batch. Ties are resolved by mean final orientation error. Action-chunk comparisons hold the model and scenes fixed to measure the effect of execution settings.

\section{Results and Analysis}
\subsection{Task 1: Bimanual Lid Opening and Return}
\begin{figure}[!t]
\centering
\includegraphics[width=\columnwidth]{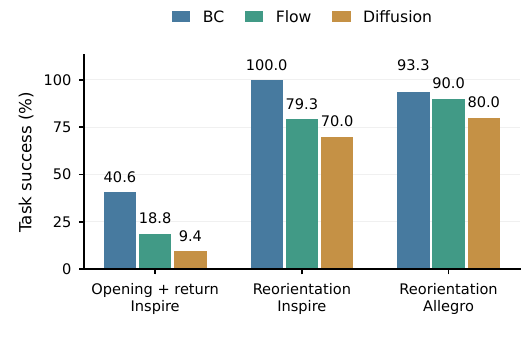}
\caption{Success rates for lid opening and return and object reorientation with two hand embodiments. Each configuration is post-trained separately and evaluated using~\eqref{eq:success}.}
\label{fig:success}
\end{figure}

The opening task starts with the Inspire hands in predefined poses. Both hands approach the box and establish contact. One primarily stabilizes the box, while the other opens the lid. They then coordinate the return motion. The complete operation lasts approximately $3.33\,$s and spans two hundred control steps. The policy must coordinate motion amplitude and phase transitions while maintaining the box's height. Figure~\ref{fig:success} reports task success rates. BC achieves $40.6\%$, while Flow and Diffusion achieve $18.8\%$ and $9.4\%$, respectively. All three methods produce complete successful trajectories. BC attains the highest completion rate. Direct regression can therefore produce executable bimanual coordination under the current demonstrations and task conditions.

\begin{figure*}[!t]
\centering
\includegraphics[width=\textwidth]{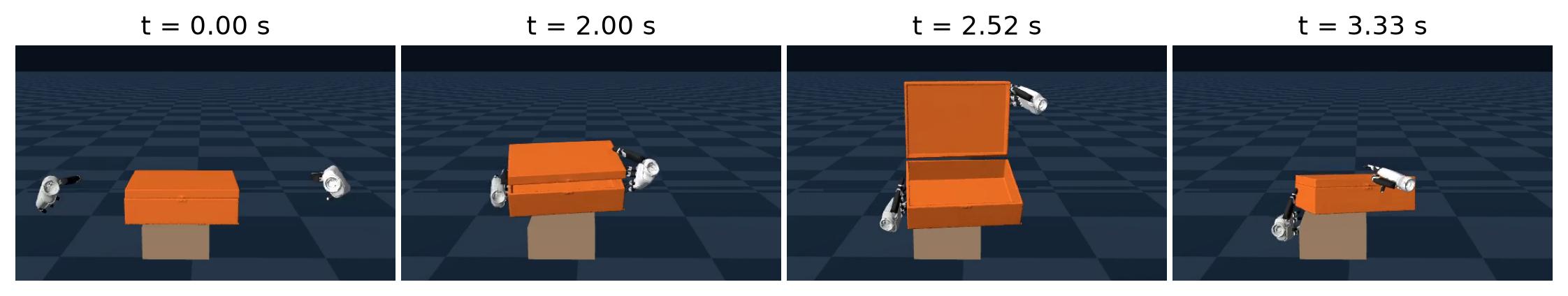}
\caption{A successful closed-loop opening and return trajectory with Inspire hands and Flow. The frontal views show the initial state, contact interaction, full opening, and the final state after lid return.}
\label{fig:opening}
\end{figure*}

Figure~\ref{fig:opening} shows a successful Flow rollout. The lid opens from its closed state into the target range and then returns. The box remains above the required height throughout the operation. As the lid angle changes, the opening hand adjusts its contact location and motion direction. The stabilizing hand also adjusts its wrist and fingers to maintain support. The trajectory therefore combines opening and return control. Compared with the short reorientation task, opening and return has a longer execution horizon and larger changes in contact relationships. Its lower success rate is consistent with these additional control demands. This comparison highlights the importance of phase transitions in continuous dexterous manipulation.

\subsection{Task 2: Bimanual Object Reorientation}
Reorientation changes the box's orientation while keeping the lid closed. The reference rotates approximately $63.96^\circ$ over $0.67\,$s. We complete expert learning, demonstration collection, and post-training for both six-channel Inspire hands and sixteen-joint Allegro hands. This evaluates the same representation and adaptation pipeline in two control spaces. On Inspire, BC, Flow, and Diffusion achieve $100.0\%$, $79.3\%$, and $70.0\%$ success. On Allegro, they achieve $93.3\%$, $90.0\%$, and $80.0\%$, respectively. BC has the highest success rate on both embodiments. Flow also achieves a high completion rate on Allegro. These results show that shared visual and temporal representations can support both active finger control spaces through embodiment-specific input and output fields.

\begin{figure*}[!t]
\centering
\includegraphics[width=\textwidth]{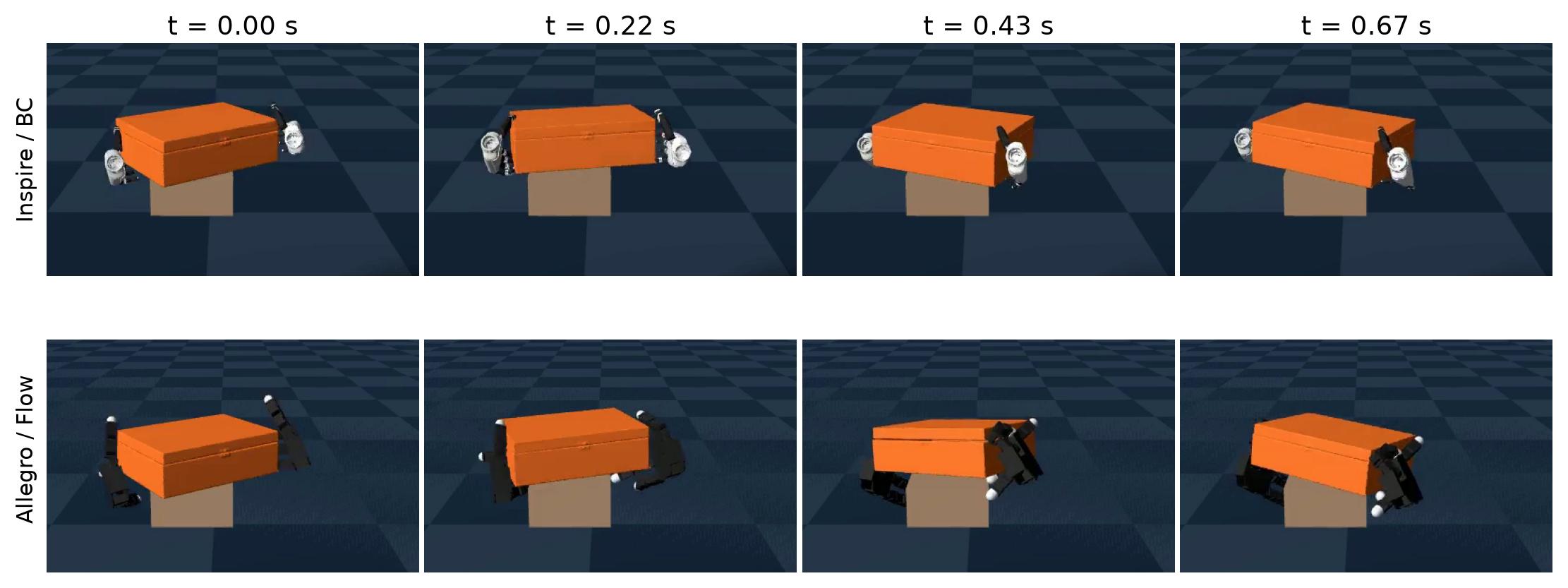}
\caption{Successful closed-loop reorientation. Top: Inspire with BC. Bottom: Allegro with Flow. Columns correspond to $0$, $0.22$, $0.43$, and $0.67\,$s.}
\label{fig:sequences}
\end{figure*}

Figure~\ref{fig:sequences} shows successful trajectories from Inspire BC and Allegro Flow. Both hands coordinate to change the box's orientation over a short interval. They maintain a closed lid and the required object height. Contact roles are relatively stable in this segment, and the main motion objective is object orientation. The model adjusts wrist and finger targets using updated images and command history. Compared with opening and return, reorientation has a shorter horizon and a more focused phase structure. All three methods achieve higher success rates on both embodiments. Together with Fig.~\ref{fig:success}, these results show that closed-loop performance varies with task structure and hand configuration. The unified interface provides a common representation and training procedure across these settings.

\subsection{Action-Chunk Continuity and Learning Behavior}
Paired comparisons show that replanning affects the continuous execution of generative policies. With fixed models in the opening task, RTC increases Flow success from $0.0\%$ to $18.8\%$ and Diffusion success from $3.1\%$ to $9.4\%$. In a separate paired Inspire reorientation comparison, Flow improves from $51.7\%$ to $79.3\%$, and Diffusion improves from $41.4\%$ to $62.1\%$. RTC constrains the overlapping region of the new plan with the remaining actions of the previous plan. The policy can therefore respond to new images while preserving temporal continuity. Command analysis also shows smaller changes in the first wrist target between plans and lower discrete acceleration. These changes accompany improved task success. Temporal alignment between generated chunks is thus an important part of converting trajectory prediction into sustained contact manipulation.

Human pretraining organizes supervision through structured wrist and finger trajectories. Robot post-training then learns action timing, motion amplitude, and embodiment-specific targets from demonstrations. Action fitting losses decrease overall as robot post-training progresses. Closed-loop evaluation measures how these predictions behave under contact dynamics and repeated replanning. Task differences and RTC comparisons link target fitting, phase coordination, and continuity between plans. AcrossDEX places these components within one pipeline. Final task success evaluates whether sequence learning produces effective physical behavior.

\section{Conclusion}
We presented AcrossDEX 1.0, a system for bimanual dexterous manipulation that connects human motion pretraining, simulation experts, robot demonstrations, and visual closed-loop control. Its central design organizes human and robot data through explicit representations and control interfaces. Human images and annotated poses support visual and motion learning. Simulation experts provide executable targets for embodiment-specific contact interaction. Shared parameters and robot post-training connect the two sources. Wrist geometry in a fixed camera frame provides shared fields. Separate finger fields and validity masks accommodate MANO poses, six-channel Inspire control, and sixteen-joint Allegro control. The visual encoder, temporal backbone, and wrist parameters are inherited from the human model. Human motion prediction and robot command prediction therefore share a defined representation and parameter transfer path.

AcrossDEX aligns temporal interfaces through physical duration. Thirty human prediction steps become sixty robot control steps, while validity masks handle complete windows and short segments. Frontal images, issued-command history, and task conditions support receding-horizon control. Future motion prediction is repeatedly updated through visual feedback. BC achieves $40.6\%$ success on opening and return, $100.0\%$ on Inspire reorientation, and $93.3\%$ on Allegro reorientation. Flow achieves $90.0\%$ on Allegro reorientation. These results demonstrate the adaptation pipeline on two hand embodiments. Differences between tasks and action-chunk comparisons also show the importance of sustained contact, phase transitions, and plan continuity. Generative policies benefit from temporal constraints inherited from the previous plan. The framework connects representation learning and target fitting to physical manipulation through a common evaluation process.

AcrossDEX establishes an implementation path from human motion data to issued robot commands. Task success measures how its components work together. Future work will extend the object set and continuous manipulation tasks, and study shared motion representations across tasks. Closed-loop interaction can support better contact stability and recovery from failures. Further experiments can examine the relationship between human data scale and robot adaptation efficiency. For physical robots, frontal vision and command history provide concrete interfaces for camera calibration, actuator modeling, and inference scheduling. By connecting data, models, and execution, AcrossDEX provides a platform for studying human motion knowledge in the control of different dexterous hands.

\bibliographystyle{IEEEtran}
\bibliography{references}
\end{document}